\documentclass[runningheads]{llncs}
\usepackage{amssymb}
\usepackage[table]{xcolor}
\usepackage[T1]{fontenc}
\usepackage {colortbl}
\usepackage{graphicx}
\usepackage{booktabs}
\usepackage{amsmath}
\begin{document}
\title{AnchDrive: Bootstrapping Diffusion Policies with Hybrid Trajectory Anchors for End-to-End Driving}
\author{Jinhao Chai \inst{1} \and
 Anqing Jiang\inst{2} \and
 Hao Jiang\inst{3} \and
    Shiyi Mu\textsuperscript{1} \and
    Zichong Gu\textsuperscript{1} \and
    Hao Sun\inst{2} \and
    Shugong Xu\textsuperscript{4*} 
    }

\authorrunning{Chai et al.}
%
\institute{
School of Communication and Information Engineering, Shanghai University, Shanghai 200444, China 
\and
 Bosch Corporate Research, Bosch (China) Investment Ltd., Shanghai, China
\and
School of Mechanical Engineering, Shanghai Jiao Tong University, Shanghai, China
\and
Xi'an Jiaotong-Liverpool University, Suzhou, China
\\
}
\maketitle              
\begin{abstract}
End-to-end multi-modal planning has become a transformative paradigm in autonomous driving, effectively addressing behavioral multi-modality and the generalization challenge in long-tail scenarios. We propose AnchDrive, a framework for end-to-end driving that effectively bootstraps a diffusion policy to mitigate the high computational cost of traditional generative models. Rather than denoising from pure noise, AnchDrive initializes its planner with a rich set of hybrid trajectory anchors. These anchors are derived from two complementary sources: a static vocabulary of general driving priors and a set of dynamic, context-aware trajectories. The dynamic trajectories are decoded in real-time by a Transformer that processes dense and sparse perceptual features. The diffusion model then learns to refine these anchors by predicting a distribution of trajectory offsets, enabling fine-grained refinement. This anchor-based bootstrapping design allows for efficient generation of diverse, high-quality trajectories. Experiments on the NAVSIM benchmark confirm that AnchDrive sets a new state-of-the-art and shows strong generalizability.
\keywords{Autonomous driving  \and Diffusion policy \and Trajectory prediction.}
\end{abstract}

\section{INTRODUCTION}
End-to-end autonomous driving algorithms have gained substantial attention in recent years owing to their superior scalability and adaptability over traditional rule-based motion planning approaches. By learning control signals directly from raw sensor data—such as camera images or LiDAR point clouds—these methods bypass the complexity of modular design pipelines, mitigate the accumulation of perception errors, and enhance overall system consistency and robustness. Earlier end-to-end planners, including UniAD\cite{uniad}, VAD\cite{vad}, and Transfuser\cite{transfuser}, relied on ego queries to regress single-modal trajectories, while more recent approaches such as SparseDrive\cite{sparsedrive} explored sparse perception modules in combination with parallel motion planners. Nevertheless, in complex traffic conditions—such as intersections or high-speed lane changes—potential vehicle behaviors can be highly ambiguous and diverse. Ignoring the inherent uncertainty in driving behavior and the multi-modal decision-making requirements imposed by environmental perception often leads to overconfident or outright failed predictions when relying on a single predicted trajectory.

Recent research has therefore begun to incorporate multi-modal modeling strategies, producing multiple trajectory proposals consistent with current scene constraints to improve decision coverage. Methods such as VADv2\cite{vadv2} and Hydra-MDP\cite{hydramdp} achieve this by using predefined discrete trajectory sets. While this increases coverage to some extent, the reliance on fixed trajectory sets inherently discretizes what is fundamentally a continuous control process, thus constraining expressiveness and flexibility.

Diffusion models have emerged as a promising alternative, offering generative and adaptive capabilities well suited for multi-modal trajectory planning. They enable direct sampling from the high-dimensional joint distribution of the ego vehicle and surrounding agents’ trajectories, and have demonstrated strong modeling capacity in high-dimensional continuous control spaces—evidenced by successes in domains such as image synthesis and robotic motion planning. Their ability to naturally model conditional distributions makes it straightforward to integrate key contextual inputs, including trajectory history, map semantics, and ego objectives, thereby improving both consistency and contextual relevance in policy generation. Moreover, their controllable test-time sampling allows for incorporating additional constraints without retraining, unlike many Transformer-based architectures.

Despite improvements such as DDIM\cite{ddim} for accelerating sampling, conventional diffusion models require numerous iterative denoising steps, resulting in high computational and latency costs at inference. To address this, prior work has shown that initializing the generation process from non-standard noise distributions can shorten the sampling path by leveraging prior information. Building on this idea, DiffusionDrive\cite{diffusiondrive} proposed a truncated diffusion strategy that anchors the process to a fixed set of trajectory anchors, enabling sampling to begin from intermediate states and thus reducing the number of required iterations. However, such fixed anchor sets lack the flexibility to adapt to scenarios demanding dynamically generated anchors.

We address this limitation with AnchDrive, a novel end-to-end multi-modal autonomous driving framework. AnchDrive employs a multi-head trajectory decoder to dynamically generate a set of dynamic trajectory anchors informed by scene perception, capturing behavioral diversity under local environmental conditions. Simultaneously, we construct a broad-coverage static anchor set from large-scale human driving data, providing cross-domain behavioral priors. These dynamic anchors provide context-aware guidance tailored to the immediate scene, while the static anchor set mitigates overfitting to training distributions and improves generalization to unseen environments. By leveraging this hybrid anchor set, our diffusion-based planner can produce high-quality and diverse predictions within a reduced number of denoising steps.

We evaluate AnchDrive in closed-loop settings on the Navsim-v2\cite{navsimv2} simulation platform, which features reactive background traffic agents and high-fidelity synthetic multi-view imagery. Experiments on a navtest set show that AnchDrive achieves 85.5 EPDMS, indicating robust and contextually appropriate behavior generation in complex driving scenarios.

Our key contributions are as follows:

\begin{itemize}
    \item We propose AnchDrive, an end-to-end autonomous driving framework that employs a truncated diffusion process initialized from a hybrid set of trajectory anchors. This approach, which integrates both dynamic and static anchors, significantly improves initial trajectory quality and enables robust planning. We validate its effectiveness on the challenging Navsim-v2\cite{navsimv2} benchmark.
    
    \item We design a hybrid perception model with dense and sparse branches. The dense branch builds a bird’s-eye-view (BEV) representation for the planner’s primary input, while the sparse branch extracts instance-level cues—such as detected obstacles, lane boundaries, centerlines, and stop lines—to enhance the planner’s understanding of obstacles and road geometry.
\end{itemize}

\section{Related Work}
\subsection{End-to-End Autonomous Driving}
Recent advances in end-to-end autonomous driving have shifted the field from traditional modular pipelines—where perception, prediction, and planning are handled independently—towards unified architectures that jointly optimize the full decision-making process. UniAD\cite{uniad} demonstrated the potential of this paradigm by integrating multiple perception tasks into a shared representation space, effectively reducing error propagation and enhancing planning performance.

Building upon this, VAD\cite{vad} introduced compact vectorized scene representations to improve computational efficiency while preserving semantic fidelity. Several subsequent works, including InterFuser\cite{interfuser} and LAV\cite{lav}, adopted a single-trajectory planning paradigm under imitation learning, where the model directly regresses a driving trajectory from raw sensor inputs. While effective in structured environments, these methods often fail to capture the inherent uncertainty and multi-modality present in complex, real-world driving scenarios. To address this, VADv2\cite{vadv2} proposed a shift to multi-modal planning by constructing a fixed vocabulary of anchor trajectories and scoring them to generate diverse outputs. Hydra-MDP\cite{hydramdp} further advanced this idea by incorporating multiple rule-based teacher policies to supervise the scoring mechanism, generating diverse candidate trajectories optimized for safety, comfort, and efficiency. This framework achieved top performance in the NAVSIM\cite{navsim} challenge, highlighting the value of integrating expert-driven evaluation into learning-based planning.

Complementary to these, Sparse methods\cite{sparsedrive}\cite{sparsemext} explored a BEV-free architecture, showing that end-to-end planning can be achieved directly from raw sensor inputs without explicit 3D reconstructions, simplifying the pipeline while maintaining competitive performance.

Recently, diffusion-based generative models have emerged as a compelling addition to end-to-end autonomous driving frameworks. Unlike regression methods that produce a single deterministic output, diffusion models generate diverse trajectories by learning to denoise from structured noise distributions. However, standard diffusion models suffer from high inference latency. To overcome this, frameworks such as DiffusionDrive\cite{diffusiondrive} and DriveSuprim\cite{drivesuprim} introduced hierarchical planning strategies: first generating coarse trajectory priors, then refining them through conditional diffusion aligned with scene semantics. These hybrid paradigms—combining candidate-based selection with generative modeling—mark a promising direction for scalable, safe, and generalizable end-to-end autonomous driving systems.

\subsection{Diffusion Models for Trajectory Prediction}

The application of diffusion models to trajectory prediction has shown significant promise in capturing multi-modal behaviors. Pioneering works in pedestrian forecasting, such as MID\cite{mid}, reframed prediction as a reverse diffusion process, while LED\cite{led} introduced acceleration techniques to learn multi-modal distributions directly. However, these models did not fully address the complexities of vehicular motion, which involves intricate interactions and adherence to traffic rules. To bridge this gap, subsequent research adapted these principles for vehicle trajectory prediction. DiffusionDrive\cite{diffusiondrive} introduced an action-conditioned denoising strategy to generate diverse, scene-consistent trajectories. Building on this, DriveSuprim\cite{drivesuprim} implemented a coarse-to-fine framework where a trajectory prior is refined via diffusion conditioned on environmental semantics and navigational intent. These hierarchical approaches enhance sampling efficiency and offer a principled way to model rich trajectory distributions, overcoming the "mean collapse" problem common in regression-based methods and providing a robust foundation for motion planning.

\section{Method}

\begin{figure}[h] 
    \centering
    \includegraphics[width=1
    \textwidth]{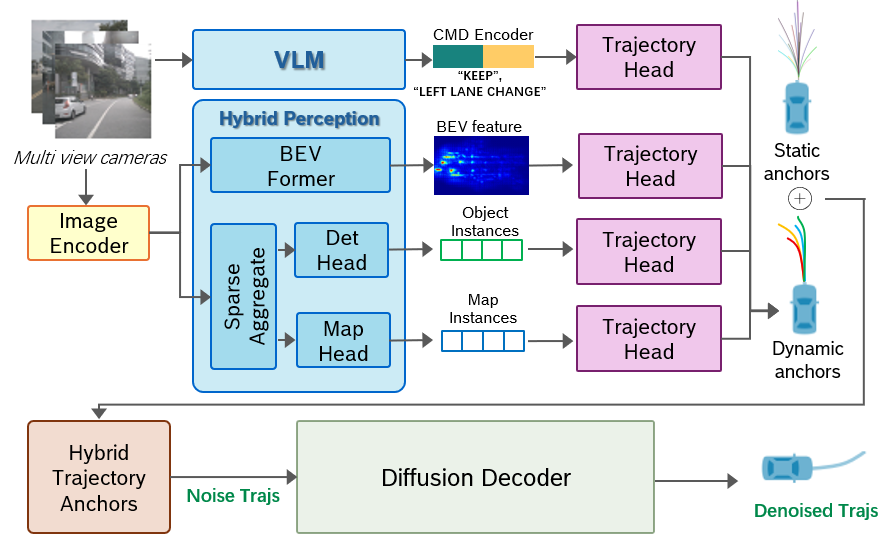} 
    \caption{The pipeline of AnchDrive. AnchDrive integrates dense- and sparse-based perception modules and employs a vision–language model (VLM) to generate high-level driving commands. The heterogeneous perception outputs and high-level commands, each carrying distinct semantic information, are processed through a multi-head attention mechanism to produce four dynamic trajectory anchors. These dynamic anchors, together with a pre-sampled static anchor set, constitute the final collection of trajectory anchors that initialize the truncated denoising process, ultimately yielding high-quality final trajectories.}
    \label{fig:pipeline} 
\end{figure}

In this section, we decompose the model architecture and analyze its components in detail. The overall architecture is depicted in Fig.\ref{fig:pipeline}.

\subsection{Hybrid Perception}

The design philosophy of our perception module is to synergistically combine implicit environmental context with explicit structural information, providing the downstream planning task with a comprehensive and robust understanding of the scene. To this end, our architecture comprises two parallel branches: a Dense Perception Branch for implicit feature extraction and a Sparse Perception Branch for explicit entity recognition.

The Dense Perception Branch aims to construct a holistic, unified representation of the surrounding environment. This branch employs a projection-based method to transform features from multi-view camera images into a single Bird's-Eye-View (BEV) perspective. We generate a 128×128 grid-based BEV feature map, covering a 64×64 meter area in the ego-vehicle's coordinate frame. This dense feature map serves as the primary contextual input, offering rich, implicit guidance on scene texture and spatial relationships to the subsequent planning module.

Complementing this, the Sparse Perception Branch is responsible for precisely extracting critical, instance-level entities from the scene. It leverages a sampling-based strategy to perform two key tasks: 3D object detection and online HD map vectorization. Its outputs are structured and explicit, including: 1) 3D object bounding boxes with attributes such as pose, size, heading, and velocity; and 2) Vectorized map elements, such as lane boundaries, road edges, and stop lines, each represented as a sequence of points. These structured outputs are subsequently encoded via an MLP to generate distinct object and map embeddings.

The core advantage of this dual-branch design lies in its synergy. While the dense BEV provides the planner with a holistic, implicit understanding of the scene, the explicit outputs from the sparse branch serve a dual purpose: they not only act as powerful feature inputs to enhance the planner's awareness of specific entities (e.g., obstacles and lanes) but can also be directly leveraged for downstream tasks such as precise collision checking and drivable area validation. This hybrid paradigm, which combines learned implicit patterns with explicit geometric constraints, overcomes the limitations of single-modality perception and provides a richer, more reliable foundation for planning decisions.

\subsection{Diffusion Policy}

Diffusion models like Denoising Diffusion Probabilistic Models (DDPMs)\cite{ddpm}, have emerged as a powerful class of deep generative models adept at capturing complex, multi-modal data distributions. This capability makes them naturally suited for modeling the inherently uncertain and multi-modal nature of autonomous driving behavior. The core principle involves two phases: a fixed forward process that gradually adds noise to data, and a learned reverse process that reconstructs the data by denoising.

In the forward process, a clean trajectory, $\tau_0$, is progressively corrupted with Gaussian noise over $T$ discrete timesteps until it becomes pure noise. The noised trajectory at any timestep $t$, denoted as $\tau_t$, can be sampled directly using the following formulation:
\begin{equation}
\label{eq:diffusion_forward}
\tau_t = \sqrt{\bar{\alpha}_t}\tau_0 + \sqrt{1 - \bar{\alpha}_t}\epsilon, \quad \text{where} \quad \epsilon \sim \mathcal{N}(0, \mathbf{I})
\end{equation}
Here, $\epsilon$ is standard Gaussian noise, and $\bar{\alpha}_t$ is a predefined noise schedule that controls the signal-to-noise ratio at each step.

The objective of the reverse process is to train a denoising network, $\epsilon_\theta$, which predicts the added noise $\epsilon$ based on the current noisy trajectory $\tau_t$, the timestep $t$, and crucially, contextual conditions $z$. In our application, this condition $z$ encapsulates information extracted from our perception module, such as Bird's-Eye-View (BEV) features, road topology, and dynamic obstacles. This transforms the denoiser into a Conditional Policy. During inference, this policy generates a trajectory by starting from an initial state and iteratively applying the learned denoising function, guided by the condition $z$, to recover a clean and feasible trajectory.

\subsection{Hybrid Trajectory Anchors for Initialized Diffusion}

A principal barrier to deploying conventional diffusion models in real-time autonomous driving lies in the substantial computational cost of the iterative denoising process initiated from pure noise. To address this limitation, we propose an initialized diffusion strategy that leverages a set of high-quality, context-aware trajectory anchors. Rather than traversing the entire denoising chain, our approach initializes the process from a curated set of anchors and executes only the final refinement stages, thereby markedly improving inference efficiency while preserving the fidelity of the generated trajectories.

This hybrid anchor set is constructed through the dynamic integration of two complementary sources:

\paragraph{Dynamic Trajectory Anchors.} To generate anchors that are both contextually relevant and aligned with the driving intent, we design a multi-head decoder that processes four heterogeneous input streams, each encoding distinct semantic information: (1) holistic Bird’s-Eye-View (BEV) scene representations; (2) sparse, object-centric features of salient instances; (3) high-definition (HD) map features capturing road topology; and (4) high-level navigation commands from a Vision–Language Model (VLM). These inputs are integrated via a multi-head attention mechanism and passed to four parallel trajectory heads, each producing a unique dynamic anchor. This architecture enables diverse intent modeling—for example, one head may prioritize adherence to a VLM-issued “turn left” command, while another focuses on avoiding a nearby obstacle.

\paragraph{Static Anchors and Hybrid Set Fusion.} To mitigate overfitting to frequently occurring training scenarios and enhance generalization to novel environments, we additionally employ a static anchor set pre-sampled from a large-scale human driving dataset. This set encodes a broad prior over diverse driving behaviors. During inference, the dynamically generated anchors are fused with this static anchor set to yield a final, comprehensive collection of trajectory anchors that jointly achieve broad coverage and high diversity.

\paragraph{Anchor-Based Trajectory Refinement.} This fused anchor set serves as the initialization for the diffusion model. Rather than synthesizing a trajectory de novo, the model predicts the residual offset between the ground truth and the closest high-quality anchor—analogous to anchor box refinement in object detection frameworks such as YOLO.

This design offers three notable advantages: (1) Efficiency — initializing from high-quality anchors substantially reduces denoising steps, meeting the stringent latency requirements of real-time planning; (2) Performance — by focusing generative capacity on fine-grained refinement instead of generation from scratch, the model achieves higher trajectory precision; (3) Robustness — the synergy between dynamic anchors (scene-specific adaptation) and static anchors (general prior coverage) markedly enhances robustness and multimodal prediction performance across complex driving scenarios.

\section{Experiments}

In this section, we first present the performance of AnchDrive on the NAVSIM v2\cite{navsimv2} navtest benchmark. Furthermore, we list the results of our ablation studies to validate the effectiveness of the proposed modules. Finally, we provide a series of visual examples to illustrate the advantages of our approach.

\subsection{Implementation Details}

We conduct experiments on the NAVSIM benchmark, a planning-oriented driving dataset derived from the OpenScene redistribution of nuPlan\cite{nuplan}. It provides $360^\circ$ coverage from 8 cameras and a fused LiDAR point cloud from 5 sensors, with annotations at 2\,Hz including HD maps and object bounding boxes. NAVSIM emphasizes challenging scenarios with dynamic driving intentions while excluding trivial stationary or constant-speed cases.

For evaluation, we adopt the Extended Predictive Driver Model Score (EPDMS) from the NAVTEST benchmark. The EPDMS aggregates multiple rule-based sub-scores, providing a holistic assessment of driving quality. It is formally defined as:
\begin{equation}
\label{eq:epdms}
\mathrm{EPDMS} = \left( \prod_{m \in \mathcal{S}_p} \mathrm{filter}m(\text{agent}, \text{human}) \right) \cdot \frac{\sum{m \in \mathcal{S}_a} w_m \cdot \mathrm{filter}m(\text{agent}, \text{human})}{\sum{m \in \mathcal{S}_a} w_m}
\end{equation}
where the $\mathrm{filter}_m$ function is used to handle cases where the human reference driver perfectly adheres to a rule, preventing potential division-by-zero issues:
\begin{equation}
\label{eq:epdms_filter}
\mathrm{filter}_m(\text{agent}, \text{human}) =
\begin{cases}
1.0 & \text{if } m(\text{human}) = 0 \\
m(\text{agent}) & \text{otherwise}
\end{cases}
\end{equation}
The sub-scores are divided into two groups. The set of multiplicative penalty scores is $\mathcal{S}_p = \{\text{NC, DAC, DDC, TLC}\}$, which correspond to \textit{no at-fault collisions}, \textit{drivable area compliance}, \textit{driving direction compliance}, and \textit{traffic light compliance}. The set of weighted-average scores is $\mathcal{S}_a = \{\text{TTC, EP, HC, LK, EC}\}$, corresponding to \textit{time-to-collision}, \textit{ego progress}, \textit{history comfort}, \textit{lane keeping}, and \textit{extended comfort}.

\label{sec:experiment}
 
\begin{table}
  \centering
  \setlength{\tabcolsep}{0.07cm} 
    \caption{Comparison on the NAVSIM navtest split based on closed-loop metrics, where higher scores correspond to better performance.}
  \label{tab:main_table}
  \begin{tabular}{@{}c|c|ccccccccc|c@{}}
    \toprule
    \textbf{Method} & \textbf{Enc.} & \textbf{NC} & \textbf{DAC} & \textbf{DDC} & \textbf{TL} & \textbf{EP} & \textbf{TTC} & \textbf{LK} & \textbf{HC} & \textbf{EC} & \textbf{EPDMS}\\
    \midrule
    Human Agent &- &100 &100 &99.8 &100 &87.4 &100 &100 &98.1 &90.1 &90.3 \\
    Ego Status MLP &- &93.1 &77.9 &92.7 &99.6 &86.0 &91.5 &89.4 &98.3 &85.4 &64.0\\
    \midrule
    Transfuser\cite{transfuser} &R34 &96.9 &89.9 &97.8 &99.7 &87.1 &95.4 &92.7 &98.3 &87.2 &76.7\\
    VADv2\cite{vadv2} &R34 &97.3 &91.7 &77.6 &92.7 &100 &99.9 &98.2 &66.0 &97.4 &76.6\\
    HydraMDP\cite{hydramdp} &R34 &97.5 &96.3 &80.1 &93.0 &100 &99.9 &98.3 &65.5 &97.4 &79.8\\
    HydraMDP++\cite{hydramdp++} &R34 &97.2 &97.5 &99.4 &99.6 &83.1 &96.5 &94.4 &98.2 &70.9 &81.4\\
    Diffusiondrive\cite{diffusiondrive} &R34 &98.0 &96.0 &99.5 &99.8 &87.7 &97.1 &97.2 &98.3 &87.6 &84.3\\
    DriveSuprim\cite{drivesuprim} &R34 &97.5 &96.5 &99.4 &99.6 &88.4 &96.6 &95.5 &98.3 &77.0 &83.1\\
    PRIX\cite{prix} &R34 &98.0 &95.6 &99.5 &99.8 &87.4 &97.2 &97.1 &98.3 &87.6 &84.2\\
    \midrule
    Diffusiondrive\cite{diffusiondrive} &V2-99 &98.2 &96.3 &99.6 &99.8 &87.5 &97.5 &97.1 &98.3 &87.7 &85.0\\
    HydraMDP++\cite{hydramdp++} &V2-99 &98.4 &98.0 &99.4 &99.8 &87.5 &97.7 &95.3 &98.3 &77.4 &85.1\\
    \rowcolor{blue!20} \textbf{AnchDrive} &V2-99 &98.0 & 97.2 & 99.6 & 99.8 & 87.2 & 97.1 & 97.7 & 98.3 & 87.9 &\textbf{85.5}\\
    \bottomrule
  \end{tabular}

\end{table}

\subsection{Quantitative Comparison}

We benchmark AnchDrive against state-of-the-art methods on the NAVSIM navtest split. Among all Camera-Only methods, AnchDrive achieves a top score of 85.5 \textit{EPDMS}, demonstrating superior performance.

Compared to VADv2\cite{vadv2}, which relies on a large, predefined vocabulary, AnchDrive shows a significant improvement of 8.9 \textit{EPDMS} while drastically reducing the number of trajectory anchors from 8,192 to just 20, a 400-fold reduction. This performance advantage extends to similar methods like Hydra-MDP\cite{hydramdp} and its variant Hydra-MDP++\cite{hydramdp++}, both of which follow the large-vocabulary sampling paradigm. AnchDrive outperforms these models by 5.7 and 4.1 \textit{EPDMS}, respectively.

Furthermore, when compared directly against DiffusionDrive (R34 backbone)\cite{diffusiondrive}, a baseline that also employs a truncated diffusion strategy, AnchDrive delivers a 1.2-point improvement in the \textit{EPDMS} metric and surpasses the baseline across all sub-scores.

Notably, these results are achieved with a fully end-to-end model learned directly from data, without relying on any hand-crafted post-processing steps.

\subsection{Ablation study}

\begin{table*}[t]
  \centering
  \setlength{\tabcolsep}{0.05cm} 
    \caption{Ablation Study on Different Trajectory Heads.}
  \label{tab:Ablation}
  \begin{tabular}{@{}cccc|ccccccccc|c@{}}
    \toprule
     \textbf{BEV} & \textbf{Obj.}& \textbf{Map.} & \textbf{VLM} & \textbf{NC} & \textbf{DAC} & \textbf{DDC} & \textbf{TL} & \textbf{EP} & \textbf{TTC} & \textbf{LK} & \textbf{HC} & \textbf{EC} & \textbf{EPDMS}\\
    \midrule
- & - & - & -                                     & 96.0 & 95.0 & 97.0 & 97.5 & 85.0 & 95.5 & 95.8 & 96.0 & 86.3 & 84.5 \\
\checkmark & - & - & -                            & 97.0 & 96.0 & 97.8 & 98.3 & 86.1 & 96.6 & 96.8 & 97.0 & 86.4 & 85.0 \\
\checkmark & \checkmark & - & -                   & 97.5 & 96.5 & 98.0 & 98.5 & 86.5 & 96.8 & 97.0 & 97.5 & 86.5 & 85.2 \\
\checkmark & \checkmark & \checkmark & -          & 97.6 & 97.0 & 99.0 & 98.5 & 86.5 & 96.9 & \textbf{97.8} & 97.7 & 87.0 & 85.3 \\
\checkmark & \checkmark & \checkmark & \checkmark & \textbf{98.0} & \textbf{97.2} & \textbf{99.6} & \textbf{99.8} & \textbf{87.2} & \textbf{97.1} & 97.7 & \textbf{98.3} & \textbf{87.9} & \textbf{85.5} \\
    \bottomrule
  \end{tabular}

\end{table*}

Table~\ref{tab:Ablation} presents our ablation study designed to systematically validate the contribution of each component within our dynamic anchor generator. We start with a baseline model that excludes all dynamic trajectory heads. Upon introducing the first head conditioned on BEV features, we observe a foundational improvement of 0.5 \textit{EPDMS}. The subsequent integration of the object feature trajectory head yields a notable increase in the \textit{NC} score, suggesting that the model effectively learns collision avoidance behaviors from the guidance of these object-centric anchors. Furthermore, incorporating the map feature trajectory head leads to gains in metrics such as \textit{DAC} and \textit{DDC}. This highlights the crucial role of explicit map awareness in enhancing the model’s adherence to road semantics and structure.

Finally, the inclusion of high-level driving commands from the VLM provides an additional boost, bringing the \textit{EPDMS} score to our final result of \textbf{85.5}.

\begin{table*}[t]
  \centering
  \setlength{\tabcolsep}{0.1cm} 
    \caption{Ablation Study on Denoising Steps.}
  \label{tab:step}
  \begin{tabular}{@{}c|ccccccccc|c@{}}
    \toprule
     \textbf{Steps.}  & \textbf{NC} & \textbf{DAC} & \textbf{DDC} & \textbf{TL} & \textbf{EP} & \textbf{TTC} & \textbf{LK} & \textbf{HC} & \textbf{EC} & \textbf{EPDMS}\\
    \midrule
    1 &97.97 & 97.15 & 99.59 & 99.79 & 87.23 & 97.13 & 97.69 & 98.30 & 87.90 & 85.43\\
    2 &97.97 & 97.21 & 99.59 & 99.78 & 87.24 & 97.14 & 97.67 & 98.30 & 87.90 & 85.48\\
    3 &98.01 & 97.19 & 99.60 & 99.79 & 87.24 & 97.13 & 97.68 & 98.30 & 87.83 & 85.49\\
    4 &98.00 & 97.18 & 99.59 & 99.78 & 87.24 & 97.14 & 97.66 & 98.31 & 87.82 & 85.46\\
    5 &97.99 & 97.17 & 99.60 & 99.79 & 87.24 & 97.10 & 97.65 & 98.31 & 87.90 & 85.46\\
    \bottomrule
  \end{tabular}

\end{table*}

We also conduct an ablation study on the number of denoising steps for our truncated diffusion model. The results, presented in Table \ref{tab:step}, reveal a key insight: for a diffusion process initialized with high-quality proposals, a greater number of denoising steps does not guarantee monotonic performance improvement. Instead, additional steps invariably increase inference latency.

Therefore, to strike an optimal balance between planning performance and computational efficiency, we select a denoising step count of 2 for the final AnchDrive model.

\subsection{Visualization}

\begin{figure}[h] 
    \centering
    \includegraphics[width=1
    \textwidth]{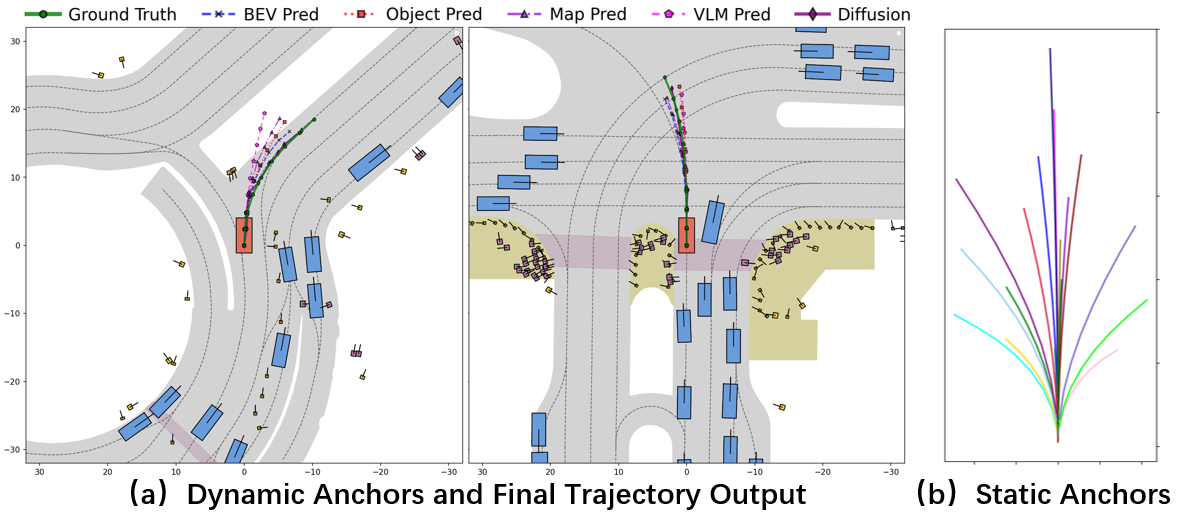} 
    \caption{Visualization in \textbf{Navtest} benchmark. (a) Dynamic trajectory anchors from our multi-head anchor generator; (b) the pre-sampled static anchor set. The two sets are fused to form the hybrid anchor set that initializes our diffusion model. The final output is shown in purple in (a), closely matching the ground truth.}
    \label{fig:vis} 
\end{figure}

To assess the generalization capability and superiority of AnchDrive, Figure \ref{fig:vis} presents qualitative results in representative driving scenarios. The right panel (b) illustrates the distribution of our static anchor set, which is derived via k-means clustering on the nuPlan\cite{nuplan} dataset and encapsulates a diverse range of common human driving maneuvers. In the two scenarios depicted in the left panel (a), our dynamic trajectory heads generate dynamic anchors that are highly relevant to the specific scene context. Notably, these dynamically generated anchors are all in close proximity to the ground-truth trajectory. This high-quality initialization enables our diffusion model to efficiently refine them into a final trajectory that aligns almost perfectly with the ground truth. These results underscore AnchDrive’s accuracy and safety in trajectory planning, particularly in challenging urban driving contexts.

\section{CONCLUSION}

In this paper, we propose AnchDrive, a novel end-to-end autonomous driving framework designed to efficiently generate diverse and safe trajectories. Its core innovation is a truncated diffusion strategy that is "hot-started" from a compact, high-quality set of hybrid trajectory anchors rather than from pure noise. This unique hybrid anchor set is formed by fusing dynamic anchors—generated from real-time scene context including BEV features, objects, maps, and VLM commands—with a static anchor set of general driving priors. This approach allows AnchDrive to leverage the powerful refinement capabilities of diffusion models. As a result, AnchDrive achieves state-of-the-art performance on the challenging NAVSIM v2 benchmark, significantly outperforming prior methods that rely on large, fixed anchor sets and demonstrating a clear advantage over similar diffusion-based techniques, thus establishing an effective and powerful new paradigm for motion planning.

\bibliographystyle{splncs04}

\end{document}